%% file: main.tex
\documentclass[letterpaper, 10 pt, conference]{ieeeconf}
\IEEEoverridecommandlockouts
\usepackage{cite}
\usepackage{amsmath,amssymb,amsfonts}
\usepackage{algorithmic}
\usepackage{graphicx}
\usepackage{textcomp}
\usepackage{xcolor}
\usepackage{url}

\def\BibTeX{{\rm B\kern-.05em{\sc i\kern-.025em b}\kern-.08em
    T\kern-.1667em\lower.7ex\hbox{E}\kern-.125emX}}
\begin{document}

\title{Design of a Compact Cable-driven 4-DoF Salamander Robotic Limb with Large Range of Motion and High Load Capacity
%a Cable-driven 4-Degree-of-Freedom Salamander Robotic Limb with High Output force and Large Workspace
}

\author{
  Zhuowen Li$^{1,2}$,
  Qiyuan Fu$^{2,*}$,
  Auke Ijspeert$^{2}$%
  \thanks{$^{1}$Shanghai Jiao Tong University, School of Mechanical Engineering}%
  \thanks{$^{2}$Biorobotics Laboratory, EPFL, 1015 Lausanne, Switzerland}%
  \thanks{$^{*}$Corresponding author: Qiyuan Fu 
  (email: qiyuan.fu@epfl.ch).}%
  \thanks{This work has been submitted to the IEEE for possible publication. Copyright may be transferred without notice, after which this version may no longer be accessible.}%
}
% \author{
%   A. Author$^{1,2}$,
%   B. Author$^{2}$,
%   C. Author$^{2}$%
%   \thanks{$^{1}$Institute 1}%
%   \thanks{$^{2}$Institute 2}%
% }

\maketitle
\input{Section/01-Abstract}
\input{Section/02-Intro-2}
\input{Section/03-Design}
\input{Section/04-Exp}
\input{Section/05-Conclusion}

\bibliographystyle{IEEEtran}
\bibliography{ref}
\end{document}

%% file: Section/01-Abstract.tex
\begin{abstract}
%Large ranges of motion (ROM) and sufficient load capacity are important for robotic limbs that aim to reproduce the diverse locomotor behaviors of amphibious quadruped animals like salamanders. The proximal hindlimb kinematics can be represented by three rotational degrees of freedom (DoFs) at the hip and one at the knee, but realizing these DoFs within a compact limb morphology is challenging because direct-drive actuation increases distal mass, limb size, and rotational inertia. 
A compact yet agile and powerful limb is essential for amphibious legged robots that transition between spine-based swimming and leg-based terrestrial walking, as salamanders do. Salamander limbs combine large ranges of motion (ROM), multiple degrees of freedom (DoFs), and high load capacity. These features enable fore–aft swinging and body-weight support during walking, as well as folding alongside the body during swimming. Their proximal kinematics can be represented by three rotational DoFs at the hip and one at the knee. Replicating these capabilities within a compact robotic limb is challenging with direct-drive architectures, because multiple joint-mounted actuators must be sufficiently spaced or placed distally to avoid self-collision, increasing limb size and rotational inertia.
Here we present a cable-driven 4-DoF salamander robotic limb with a spatial cable-routing architecture that relocates actuators proximally while preserving large ROM and load capacity. Transmission-induced coupling is mitigated by mechanically compensating multi-segment cable paths, model-based feedforward decoupling, and output-side joint angle feedback. Compared with a direct-drive limb, the proposed design reduces mass by 20.4\%, projected envelope area by 20.5\%, and estimated effective yaw-axis inertia by 31.9\%. Experiments verify independent, decoupled control over joint-angle ranges of \(45^\circ\), \(118^\circ\), \(118^\circ\), and \(135^\circ\) for hip pitch, yaw, roll, and knee flexion/extension, respectively. During lifting tests, the limb supported an equivalent vertical load of at least 5.4 times its own weight. The proposed limb provides a compact, biologically representative platform for future studies of salamander locomotion and neural control.
%achieves a ROM (1.32 $L^3$, where L denotes the limb length) sufficient for locomotion and can lift an external load approximately 6.7 times its own mass. The proposed limb provides a biologically representative platform for future studies of salamander locomotion and neural control.
\end{abstract}

% \begin{IEEEkeywords}
% component, formatting, style, styling, insert.
% \end{IEEEkeywords}

%% file: Section/02-Intro-2.tex
\section{Introduction}

% Salamanders have attracted considerable attention in robotics and neuroscience because of their remarkable locomotor abilities in both aquatic and terrestrial environments \cite{Auke07science}. Coordinated motions of the body, tail, and limbs enable versatile locomotion across different terrains, including walking, paddling, and swimming.\cite{understanding_salamander_locomotion} To investigate how neural mechanisms generate these movements \cite{auke_neurosciences}, several salamander-inspired robotic platforms have been developed as scientific tools for testing biological hypotheses\cite{liuCPG}\cite{Qiyuan26ICRA}\cite{Auke13TRO}. However, many existing salamander robots simplify each limb to one or two active degrees of freedom, which supports basic locomotor behaviors but cannot reproduce the kinematic capacity of salamander limbs, thereby limiting studies of neural control involving richer and more biologically relevant movements \cite{Auke13TRO}\cite{Pleurobot}\cite{auke_amphibious}.

Owing to their foldable limb morphology as well as the flexible spine, salamander robots are promising platforms for combining agile legged locomotion on land with efficient, spine-dominated swimming in fluids \cite{Auke13TRO, Pleurobot}. However, whereas salamanders rely on coordinated, complex motions of both the limbs and body for multimodal locomotion \cite{understanding_salamander_locomotion}, most existing salamander robots \cite{auke_amphibious} simplify each limb to only one or two degrees of freedom (DoFs). Although sufficient for basic walking, these reduced limb designs cannot reproduce the full kinematic capacity of salamander limbs, limiting the locomotion capacity of salamander robots in complex environments \cite{Pleurobot, auke_amphibious}.

\begin{figure}[th]
  \centering
  \includegraphics[scale=0.43]{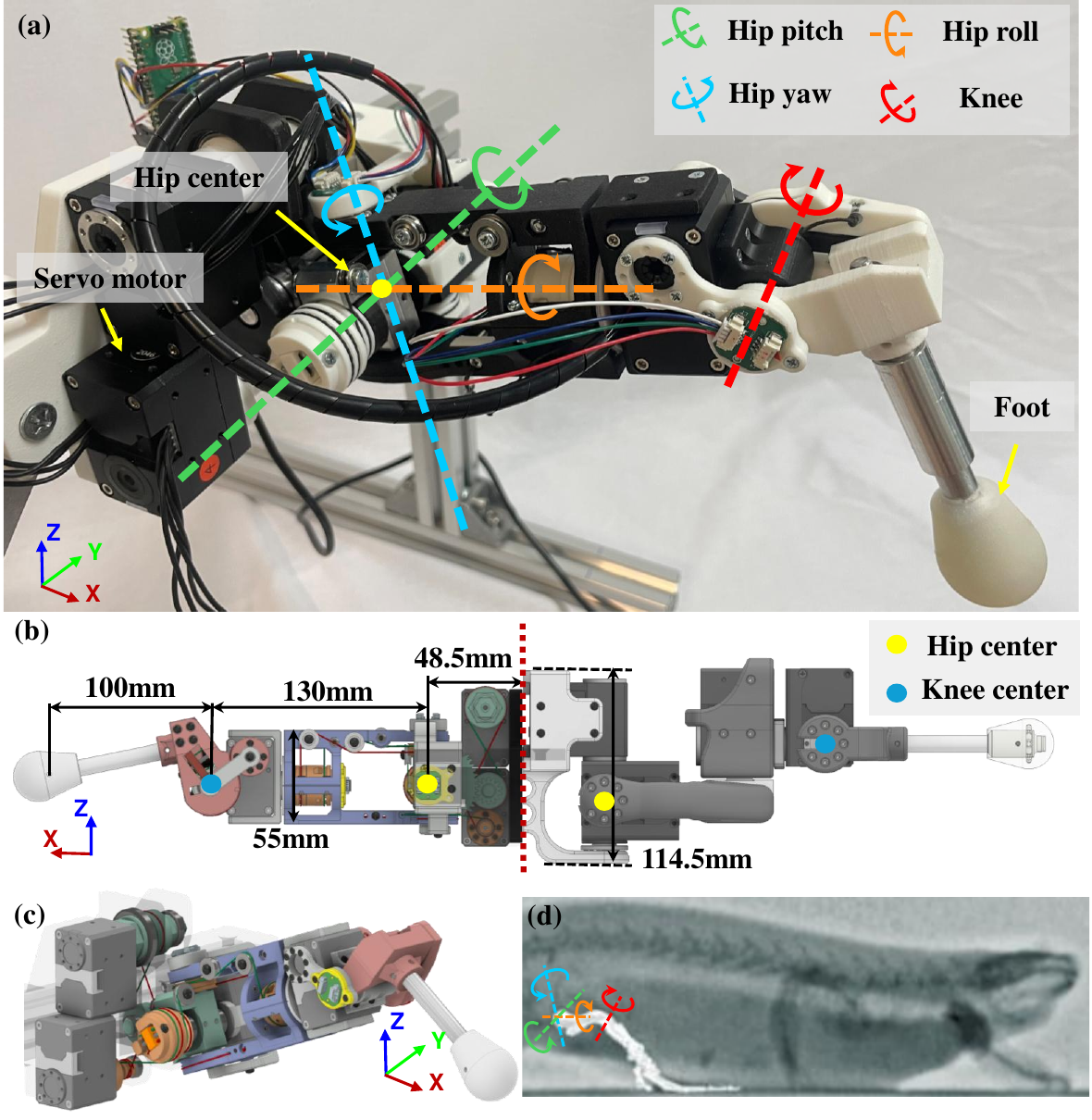}
  \vspace{0cm}
  \caption{Overview of the cable-driven 4-DoF salamander robotic limb. (a) Overview with three mutually orthogonal hip axes intersecting at the hip center and a separate knee flexion/extension axis. (b) Principal dimensions of the proposed cable-driven limb (left) compared with the previous direct-drive limb (right). (c) CAD model showing actuator placement and spatial cable routing. (d) Salamander right hind limb’s bones in the side view of a sequence of X-ray snapshots \cite{pleurobot_phd_thesis}. Joint-specific transmission components are color-coded in sage green for pitch, indigo for yaw, orange for roll, and terracotta for knee. Red and green lines indicate the antagonistic cables, whereas motors and non-joint-specific components are shown in neutral colors.}
  \label{fig:overview}
  \vspace{0cm}
\end{figure}
To effectively reproduce the versatile multi-modal locomotion of salamanders, salamander-inspired robotic hindlimbs should retain four principal DoFs: three mutually orthogonal rotational degrees of freedom at the hip and one flexion-extension degree of freedom at the knee \cite{understanding_salamander_locomotion}. The three hip DoFs act together to generate three-dimensional limb retraction and positioning, contributing to propulsion, stride length, and adaptation to uneven terrain, while knee flexion–extension modulates effective limb length and provides foot clearance during swing. Moreover, the three hip joint axes intersect at a single point, providing a closer anatomical abstraction of the salamander hip \cite{pleurobot_phd_thesis}.

Some previous salamander robots \cite{Pleurobot,liuCPG} incorporated a 3-DoF hip to better approximate salamander limb kinematics. However, the limb design adopted a direct-drive configuration in which each geared servo is mounted directly at its corresponding joint, without an additional external transmission. Although this simplified the mechanical design, the available joint torque was limited by the servo output, while the hip motors had to be spaced sufficiently apart to avoid self-collision, increasing limb rotational inertia and projected area. % achieve biological fidelity, compact morphology

These limitations motivate a transmission system that enables actuator relocation toward the trunk while providing the required joint torque. Several transmission mechanisms have been investigated for robotic limbs \cite{leg_review}.
Linkages offer stable load transmission \cite{haoqileg}, but they either require substantial space for link rotation or make it difficult to achieve large ranges of motion without encountering dead points. Gears provide reliable short-distance transmission \cite{differential_dog_leg} but add a significant mass when driving all four DoFs. Belt drives \cite{Stanford_doggo} transmit torque over longer distances, but their planar routing and limited torque capacity at the required scale conflict with the limb's design requirements.

In contrast, cable-driven mechanisms offer a powerful yet flexible transmission approach that can transmit large torques over long distances. Researchers have developed high-performance cable-driven robotic arms and dexterous hands\cite{Cable_Arm_Kim}\cite{TwistSnake}\cite{Hand_haoqi}, as well as robotic legs with impressive dynamic locomotion capabilities\cite{cable_driven_dog_leg}. 

% However, there are two critical problem needed to be solved in cable-driven systems, which are pretension and decouple.Pretension mechansim is an essential component in cable-driven systems to eliminate backlash and dead zones. The basic principle of the pretension mechanism is to use unidirectional transmission to extend the cable routing through component rotation or translation and utilize self-locking characteristics to balance with the cable tension.

% Pretension mechanism is an essential component in cable-driven systems to eliminate backlash and dead zones. 

Despite these advantages, compact multi-DoF cable-driven mechanisms require effective cable pretensioning. Pretensioning mechanisms typically rely on unidirectional transmission to increase cable tension by extending the cable path. Various studies have employed one-way bearings\cite{TITAN-XIII}, clutches\cite{Low_Cost_Cable-Driven_Robot_Arm}, and worm gears\cite{Birdbot} for pretensioning. However, these mechanisms require high manufacturing precision that is hard to achieve for customized compact components using 3D printing. Spring-based tensioning methods have also been widely utilized in dexterous hands \cite{Hand_haoqi}\cite{hand_spring}, but under high-load conditions, high-stiffness springs may result in bulky components.

Another challenge in using cable-driven mechanisms is to decouple the four non-planar DoFs. Many cable-decoupling methods are designed for planar mechanisms, including rolling contact joints\cite{Cable_Arm_Kim, decouple_Geared_Rolling_Joints, D3-arm}, four-bar linkages\cite{decoupling_four_link}, and movable pulleys \cite{decouple_movable_pulley}. Adapting these mechanisms to the present intersecting-axis hip would require additional routing or decoupling components, increasing the packaging volume. Three-dimensional mechanical decoupling solutions have also been developed \cite{Constant_Length_Tendon,decoupling_rotary_pulley,D3-arm}, but they rely on separate decoupling assemblies that occupy additional volume rather than being integrated into the joint structure. This packaging overhead conflicts with the compact three-axis hip required here.

In this work, we present a compact cable-driven 4-DoF robotic limb that preserves the biological joint configuration through spatial cable routing. To preserve compactness, screw- and ratchet-based \cite{ORCA_Hand} pretensioning mechanisms are integrated into load-bearing components. To mitigate transmission errors arising from DoF coupling and long cable paths, we developed mechanically compensating cable routing to suppress variations in total cable length, geometric feedforward decoupling to compensate for differential-length coupling, and output-side feedback to correct residual tracking errors.

Compared with a direct-drive limb, the proposed limb reduces mass by 20.4\%, projected envelope area by 20.5\%, and estimated effective yaw-axis rotational inertia by 31.9\%. 
Experiments demonstrate that it preserves a large ROM while suppressing transmission-induced errors and maintaining high load capacity. Together, these results demonstrate the feasibility of integrating the proposed limb into a robotic platform for future studies of multimodal amphibious locomotion.

% Experiments further demonstrate that the compact limb preserves a large ROM, suppresses transmission-induced errors, and provides high load capacity.
% The measured foot-position spans along the \(x\)-, \(y\)-, and \(z\)-axes are 78.9\%, 164.3\%, and 111.3\% of \(L\), respectively, where $L$ denotes the limb length from the hip center to the foot tip. The CAD-derived theoretical workspace has a volume of 1.28 $L^3$, although full experimental validation remains pending because of current software constraints.
% The roll joint, despite having the longest transmission path, exhibits a mean absolute angle error of \(0.91^\circ\). The mean three-dimensional foot-position error is 2.08\% of \(L\). 
% %\(16.02 \times 10^{6}\,\mathrm{mm}^{3}\)  %The maximum joint-angle and foot-position deviations are $4.88^\circ$ and 8.2~mm, respectively.
% The prototype cab also lift an external load equivalent to 5.4 times its own weight. 
% Together, these results demonstrate the feasibility of integrating the proposed limb into a robotic platform for future studies of multi-modal amphibious locomotion.

%% file: Section/03-Design.tex
\section{Design}
\subsection{Biological reference}
Salamanders rely on their limbs for diverse terrestrial and aquatic behaviors, including turning, paddling, and folding along the trunk to reduce hydrodynamic drag. Based on the anatomical structure of \textit{Pleurodeles waltl}, a salamander hindlimb can be represented as a four-DoF chain: three rotational DoFs at the hip—yaw, pitch, and roll about three mutually orthogonal hip axes that intersect at a single point—and knee flexion/extension \cite{Pleurobot}. This 4-DoF chain is kinematically redundant when only foot position is considered, but the hip-roll DoF is found to also contribute to body propulsion during the final one-third of the stance phase in salamander walking \cite{understanding_salamander_locomotion} and reorient the distal limb for additional control over limb posture. 
%Thus, although omitting hip roll would substantially simplify the mechanical design, it would reduce postural flexibility in complex amphibious environments. 
Therefore, the proposed limb retains the biological 4-DoF joint configuration and the ranges of motion described in \cite{Pleurobot}.

\subsection{Overview}
% Table 1 summarizes its main specifications. 
Based on the demands for salamander-like locomotion in existing studies \cite{Pleurobot,understanding_salamander_locomotion}, we developed the cable-driven robotic limb shown in Fig. \ref{fig:overview} (a). The limb geometry adopts an 8.6:1 scale relative to \textit{Pleurodeles waltl}, resulting in an overall limb length \(L\) of 230~mm. The assembled prototype has a mass of 794~g. The direct-drive limb used for comparison was constructed following \cite{Pleurobot}, with an additional force sensor of mass 100~g located 81~mm from the yaw axis. Excluding the sensor, the limb has a mass of 997~g and the same upper- and lower-link lengths as the cable-driven limb. The envelope areas of the limb profiles projected onto the limb-longitudinal-vertical plane in Fig.~\ref{fig:overview} (b) are $0.318L^2$ and $0.400L^2$ for the cable-driven and direct-drive limbs, respectively, representing a 20.5\% reduction.

To match salamander walking dynamics at this scale, the target locomotion frequency is 0.2 Hz. The limb must support up to 40\% of the weight of an approximately 8-kg robot during stance \cite{Pleurobot, kawano_blob_2022}. These requirements correspond to a maximum pitch-joint torque of approximately 7~N·m.

To reduce the rotational inertia, the proposed design places all hip motors proximally. The hip pitch, yaw, and roll joints, followed by the knee joint, are arranged sequentially from proximal to distal. All four joints are driven by antagonistic cable pairs, with the cable ends terminated with Ashley stopper knots. Ratchet- and screw-based cable pretension mechanisms are incorporated into load-bearing components, avoiding separate tensioning assemblies and preserving the compact joint envelope. A 0.8-mm Dyneema cable was selected to accommodate small routing radii while providing high tensile strength and low extension.

Hall-effect sensors (Infineon Technologies AS5048B) measure the output angles for direct joint-angle feedback. 
Most structural components are 3D-printed in polylactic acid (PLA) or carbon-fiber-reinforced nylon (Markforged Onyx), while the load-bearing central cross-shaft is CNC-machined from aluminum to provide the required strength and dimensional accuracy.

% \begin{table}[t]
%   \caption{specifications}
%   \label{tab:limb-specifications}
%   \centering
%   \footnotesize
%   \renewcommand{\arraystretch}{1.08}
%   \setlength{\tabcolsep}{4pt}
%   \begin{tabular}{@{}p{0.42\columnwidth}p{0.50\columnwidth}@{}}
%     \hline
%     \textbf{Item} & \textbf{Specification} \\
%     \hline
%     Motor & Dynamixel XM430-350 (ROBOTIS) \\
%     Cable & Dyneema \\
%     Hall sensor & AS5048B \\
%     \hline
%   \end{tabular}
% \end{table}

\subsection{Joint Mechanism}

\begin{figure}[th]
  \centering
  \includegraphics[scale=0.35]{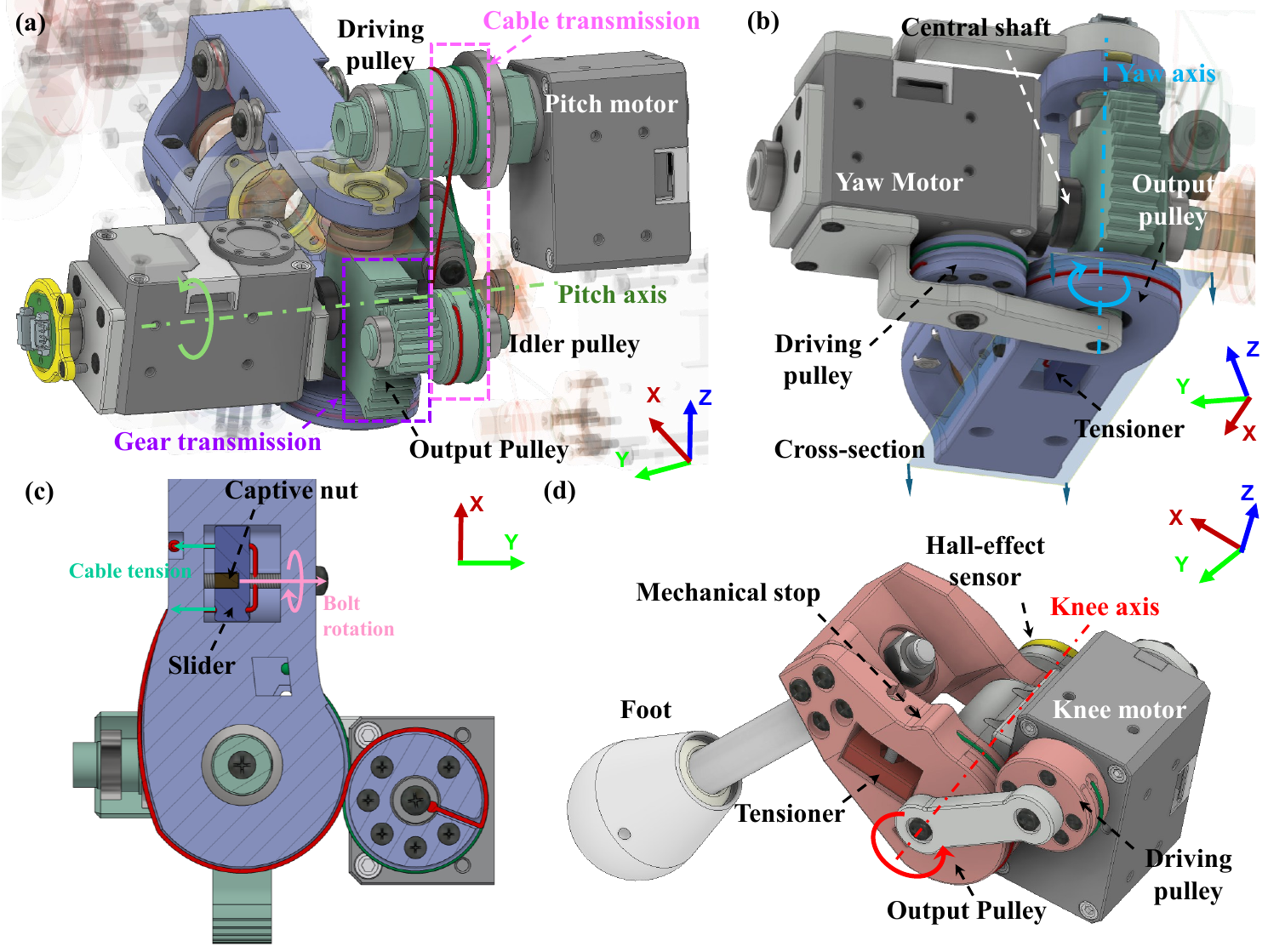}
  \vspace{0cm}
  \caption{Cable transmissions and pretension mechanisms for (a) hip pitch with two-stage reduction mechanism, (b-c) hip yaw, and (d) knee flexion/extension. (c) is a cross-sectional view along the blue plane indicated in (b). Cable tensioning is realized by a ratchet mechanism in (a) and a screw-adjusted sliding pretensioner in (b-d).  Red and green curves denote the antagonistic cables.}
  \label{fig:other joint mechanism}
  \vspace{0cm}
\end{figure}

Within the common cable-driven architecture, the reduction, decoupling, and pretensioning mechanisms are tailored to the torque, routing, and packaging constraints of each joint. Candidate transmission ratios were evaluated using the joint torque and angular-velocity requirements measured from representative salamander locomotion gaits using the direct-drive limb equipped with Dynamixel XM540 servo motors, together with the torque-speed characteristics of candidate servo motors. The final ratios were selected to meet these requirements while preserving the required range of motion and fitting the pulley layout within the available limb volume. The selected ratios are \(4.22\), \(1.50\), \(2.79\), and \(1.50\) for hip pitch, yaw, roll, and knee, respectively. These reduction ratios enabled the use of smaller Dynamixel XM430 servo motors.

\subsubsection{Hip pitch}

\begin{figure}[th]
  \centering
  \vspace*{2mm}
  \includegraphics[scale=0.32]{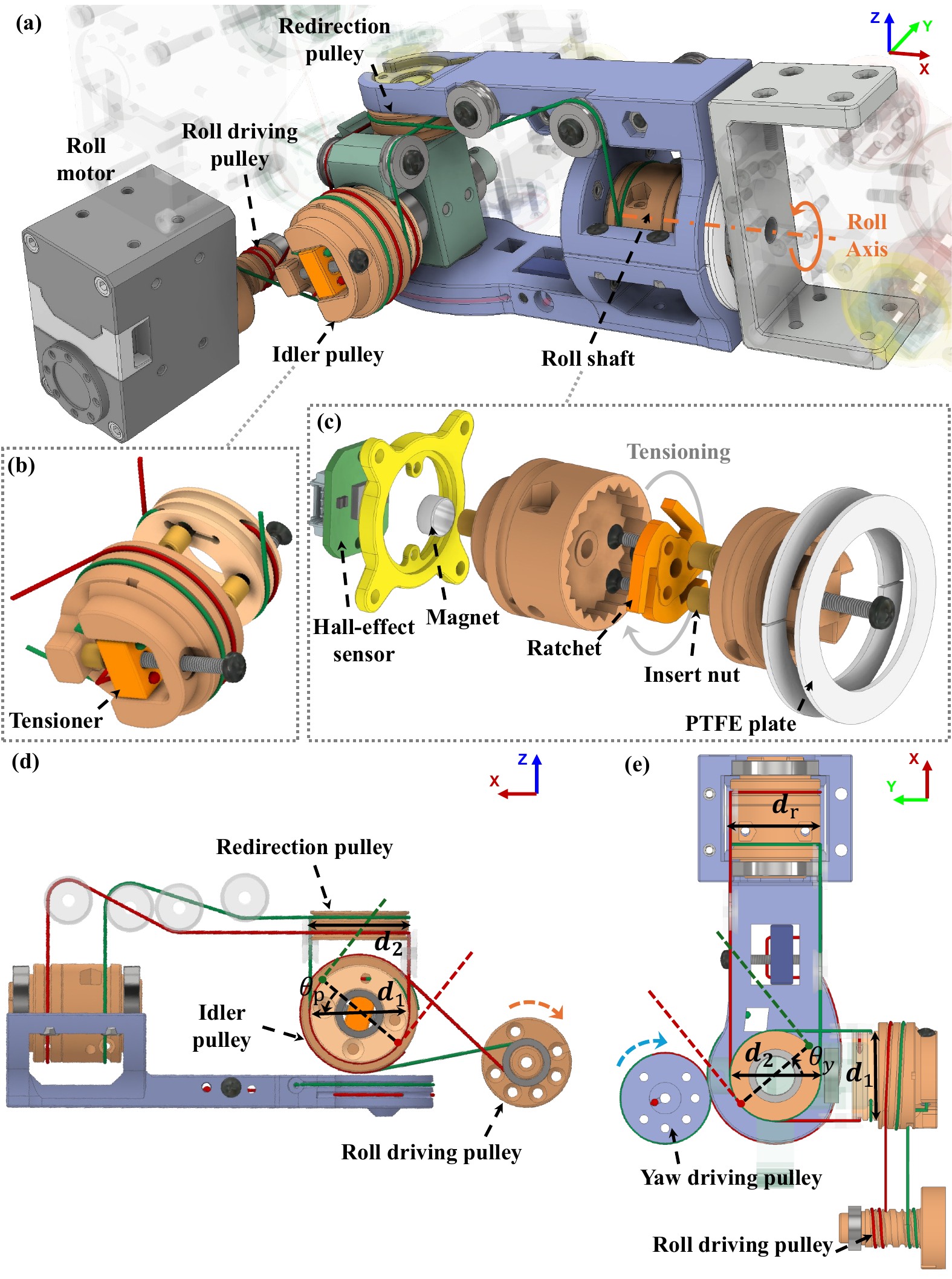}
  \vspace{0cm}
  \caption{Hip-roll transmission and decoupling geometry. (a) Overview showing cable routing from the roll motor to the roll shaft. (b) Exploded view of the idler pulley with a screw-based tensioner. (c) Exploded roll-shaft assembly showing the ratchet-based tensioner, PTFE thrust plates, and coaxially aligned magnet and Hall-effect sensor. (d-e) Cable-routing geometry illustrating pitch-induced (d) and yaw-induced (e) differential-length changes used to derive the feedforward compensation. The pulley diameters $d_1$ and $d_2$, roll-shaft diameter $d_r$, and joint angles $\theta_p$ and $\theta_y$ define the geometry used for model-based feedforward decoupling. Red and green curves denote the antagonistic cables. Dashed curves illustrate cable motion with the coaxial shaft, during which equal and opposite cable-length compensation occurs.}
  \label{fig:roll joint mechanism}
  \vspace{0cm}
\end{figure}

% The pitch joint has the highest torque requirement and need to make space for motors and transmission mechanisms for other distal joints. Therefore, it uses a compact two-stage reduction combining cable and gear stages (Fig. \ref{fig:other joint mechanism} (a)). 
The pitch joint has the highest torque requirement and needs to make space for motors and transmission mechanisms of other distal joints. Therefore, it employs a two-stage reduction to avoid using a large output pulley (Fig. \ref{fig:other joint mechanism}(a)).
In the first stage, the trunk-mounted motor drives the larger pulley of a compound idler through cable transmission. A ratchet integrated into the pitch driving pulley provides cable pretensioning. The same mechanism is described in detail below for the roll joint (Fig. \ref{fig:roll joint mechanism} (c)).
In the second stage, to facilitate assembly and maintenance in limited space, the smaller coaxial pulley on the idler drives the pitch output through a gear pair.

\subsubsection{Hip yaw}

The yaw motor is mounted on the central shaft and rotates with the pitch assembly. The yaw output uses the rolling contact joint \cite{rolling_hand} shown in Fig. \ref{fig:other joint mechanism} (b). Its screw-based tensioner routes one cable through two oppositely directed holes in a movable cube containing a captive nut. Tightening the bolt translates the cube along the screw axis against the cable-tension direction, increasing the effective cable-path length and applying pretension, as illustrated in Fig. \ref{fig:other joint mechanism} (c). Notably, this path-length adjustment is required for only one cable because the two antagonistic cables are coupled through the driving and joint pulleys.

\subsubsection{Hip roll}

When transmitting the torque from the trunk-mounted roll motor to the roll shaft, the intermediate hip pitch and yaw DoFs induce cable-length changes that must be compensated. These changes comprise two components: (1) simultaneous extension or retraction of both cables due to changes in the 3-D distances between pulley centers, which changes the total length; and (2) extension of one cable accompanied by retraction of the other, which changes the differential length.

We compensate for differential-length changes in software (Section \ref{sec:decoupling}) and total-length changes mechanically using a two-segment cable transmission (Fig. \ref{fig:roll joint mechanism} (a)). Mechanical compensation is achieved primarily by an idler pulley coaxial with the pitch shaft and a redirection pulley coaxial with the yaw shaft, both of which rotate freely about their respective shafts. 

The fixed distances between successive pulleys eliminate total-length changes, allowing the antagonistic cable pair to undergo purely differential-length changes as the coaxial shafts rotate, thereby preventing cable slack or excessive tension.

The first segment connects the driving pulley to the idler pulley. The helical grooves on these two pulleys separate successive cable turns and prevent stacking as the driving pulley rotates by more than one revolution, and their pitches are matched to the reduction ratio between the pulleys. Given the limited thickness of the idler pulley, this segment uses the screw-based tensioner shown in Fig. \ref{fig:roll joint mechanism} (b). 

The second segment continues from the idler pulley through a redirection pulley to the roll shaft, whose exploded view is shown in Fig. \ref{fig:roll joint mechanism} (c). The ratchet integrated between the two roll-shaft components permits relative rotation in the tensioning direction, while its pawls prevent reverse rotation under cable tension, thereby maintaining cable pretension. A through-bolt connects the assembly and increases its bending stiffness, while PTFE plates act as thrust bearings to carry the compressive forces generated by the bending moment.

\subsubsection{Knee}

Finally, the knee motor is mounted on the distal end of the roll shaft. As shown in Fig. \ref{fig:other joint mechanism} (d), the knee joint adopts the same rolling contact joint mechanism for reduction and screw-based tensioner as the yaw joint.
The knee joint axis is parallel to the hip pitch axis when the hip yaw and roll angles are both zero. A mechanical stop integrated into the knee output pulley provides a repeatable reference for joint-angle calibration.

\subsection{Sensory feedback and calibration}

At each joint output, an embedded magnet rotates with the joint, and a coaxially aligned Hall-effect sensor measures its orientation for direct output-side joint-angle feedback. 

% The joint-angle sensors were calibrated and their outputs were checked before limb assembly.  \TODO{ADD DETAILS} 

After limb assembly and cable pretensioning, the joint-angle zero offsets were calibrated using a Vicon motion-capture system. Two rigid marker frames were used: one was fixed to the trunk, and the other was mounted in place of the ball foot. Based on the measured relative orientation, the servo-motor offsets were adjusted to bring the limb to its predefined reference configuration. The corresponding Hall-sensor readings were then recorded as the joint-angle zero references.

\subsection{Feedforward decoupling and feedback control} \label{sec:decoupling}

Pitch and yaw rotations alter the differential path lengths of the antagonistic roll-cable pairs, thereby inducing unintended roll motion. Although direct joint-angle feedback can correct the resulting error, we additionally develop a geometric feedforward compensation to decouple this motion.

The routing maintains constant lengths for the cable portions not wrapped around the idler pulley or the redirecting pulley, so coupling arises only from the wrapped portions. Taking yaw rotation as an example, the cable-pulley tangent points move along the circumference of the redirection pulley as the yaw joint rotates, producing equal and opposite length changes in the two antagonistic cables, as shown in Fig. \ref{fig:roll joint mechanism} (e). Similarly, pitch rotation produces the corresponding cable-length variation at the idler pulley. Therefore, the roll-shaft rotation induced by pitch and yaw rotations is

\begin{equation}
\Delta\theta_r=\frac{-d_1\theta_p+d_2\theta_y}{d_r}
\label{eq:ff_compensation}
\end{equation}

where $\theta_p$ and $\theta_y$ are the pitch and yaw angles, respectively, and $d_1=d_2=d_r=24~\mathrm{mm}$ are the diameters of the idler pulley, redirection pulley, and the roll shaft. The controller computes $-\Delta\theta_r$ in real time and commands the corresponding motor-side rotation. 

After most of the coupling error is compensated by this feedforward term, direct joint-angle feedback corrects the residual error at 50 Hz. In the following experiments, this feedback controller was used only in the roll and yaw load tests. The open-loop control used in the other tests demonstrates the effectiveness of the proposed mechanical design. For the roll joint, however, the longer transmission path resulted in larger errors under high loads, making feedback control necessary.

\begin{equation}
\begin{aligned}
e_r(t) &= \theta_{r,\mathrm{tar}}(t)-\theta_r(t),\\
I_r(t) &= \operatorname{sat}_{I_{\max}}
\left(\int_0^t e_r(\tau)\,\mathrm{d}\tau\right),\\
\theta_{r,\mathrm{cmd}}(t)
&= \theta_{r,\mathrm{tar}}(t)+\theta_{\mathrm{ff}}(t)
+K_P e_r(t)+K_I I_r(t).
\end{aligned}
\label{eq:pi_control}
\end{equation}

where $\theta_{r,\mathrm{tar}}$ is the target roll angle, $\theta_{r,\mathrm{cmd}}$ is the motor command expressed in joint-angle coordinates, $\theta_r$ is the roll angle measured by the Hall-effect sensor, $\theta_{\mathrm{ff}}=-\Delta\theta_r$ is the feedforward compensation angle, and $I_r$ is the roll-angle error integral subject to the limit $I_{\max}$. $K_P=2.0$ and $K_I=1.0$ are the proportional and integral gains. For the other joints, \(\theta_{\mathrm{ff}}=0\). 

%% file: Section/04-Exp.tex
\section{Experiment}

\begin{figure}[th]
  \centering
  \vspace*{3mm}
  \includegraphics[scale=0.32]{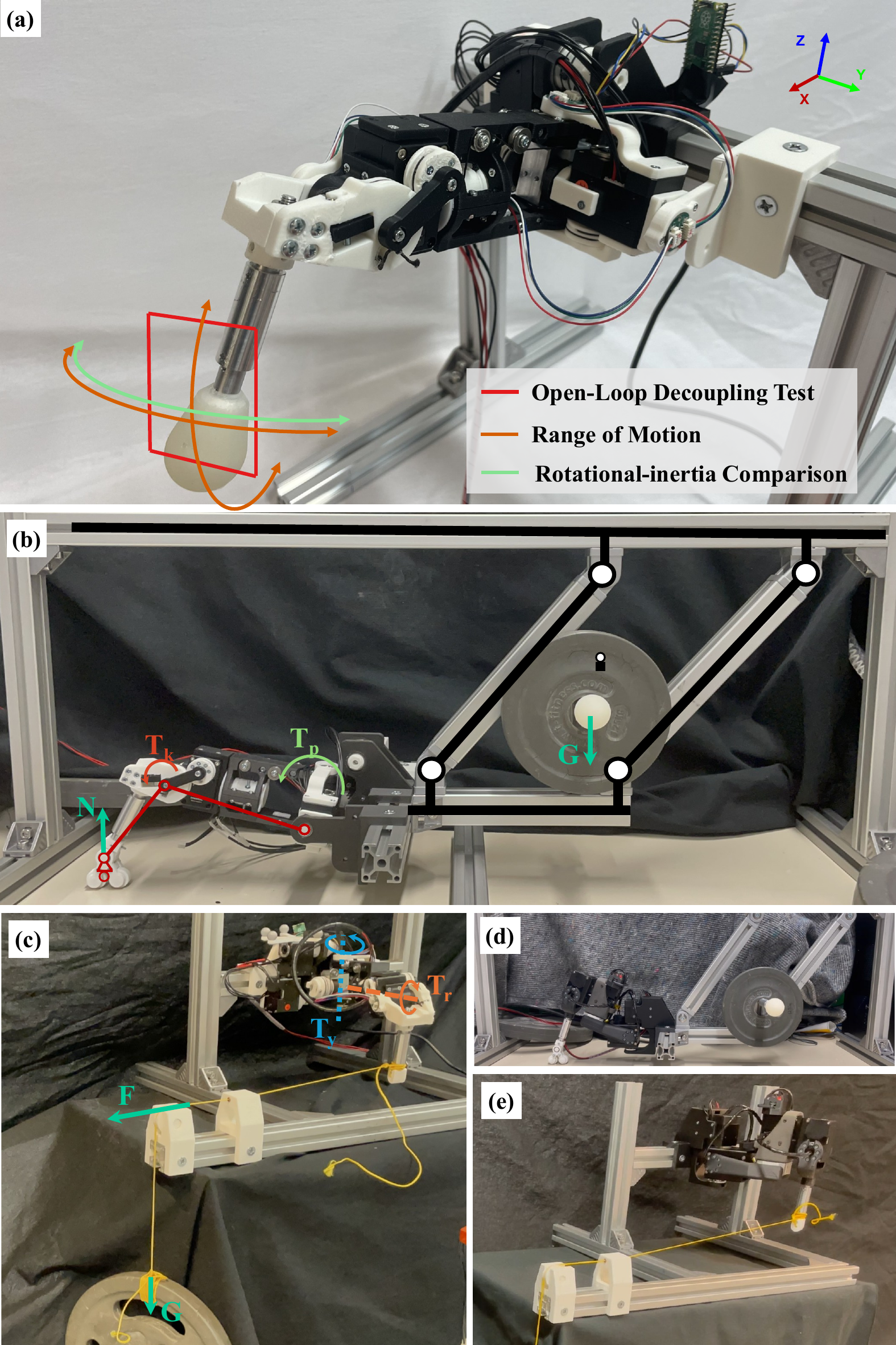}
  \vspace{0cm}
  \caption{Experimental setup. (a) Setup for the decoupling, workspace, and inertia tests with target motions annotated. (b) Cable-driven limb loaded through a four-bar linkage to evaluate pitch and knee joint load capacity. (c) Cable-driven limb subjected to a horizontal cable force to evaluate the yaw and roll joint load capacity. (d, e) Direct-drive limb tested using the corresponding pitch/knee and yaw/roll loading setups, respectively. Here, $G$ denotes the gravitational force, $N$ the vertical foot reaction, $F$ the horizontal cable force, and $T_p$, $T_k$, $T_y$, and $T_r$ the pitch, knee, yaw, and roll joint torques, respectively.}
  \label{fig:experiment_platforms}
  \vspace{0cm}
\end{figure}

\subsection{Open-Loop Decoupling Test}

To evaluate the effectiveness of the feedforward decoupling method, we compared foot-tip trajectories during tracking of a square trajectory using inverse kinematics with and without feedforward decoupling. Output-side joint-angle feedback was disabled and the target roll angle was set at zero position in both trials. 

Feedforward decoupling suppressed the unintended roll motion induced by yaw and pitch rotation (Fig. \ref{fig:decouple_inertia}(a), orange dotted), reducing the mean three-dimensional foot-position tracking error from 11.8 mm to 9.2 mm (Fig. \ref{fig:decouple_inertia}(b)). A substantial portion of the absolute error was attributable to a constant offset between the experimental and theoretical coordinate origins. After translation-only alignment using the position of the starting point, the mean tracking errors decreased to 5.7 mm and 2.5 mm for the undecoupled and decoupled cases, respectively.

\subsection{Rotational-inertia Comparison} \label{sec:exp_inertia}

To compare the effective rotational inertia of the two limbs about the yaw axis, we measured the yaw-motor torques and angular velocity when the yaw joints were commanded to track the same repeated sinusoidal trajectory, spanning $-75^\circ$ to $40^\circ$ with a period of 1.5 s (\(0.67~\mathrm{Hz}\)). For a fair comparison with the direct-drive limb, the measured motor torque of the cable-driven limb was scaled by the cable-transmission reduction ratio to obtain the corresponding joint-side torque. The motor torque was estimated from the measured motor current assuming a linear current-to-torque relationship \cite{xm430w350emanual}, sampled at 50 Hz. The joint angular acceleration was estimated as the discrete derivative of the joint-side angular velocity, obtained by dividing the angular velocity read from the motor by the reduction ratio. The joint acceleration and torque signals were filtered using identical zero-phase Gaussian low-pass filters (\(\sigma=0.10~\mathrm{s}\), approximate cutoff frequency: \(1.33~\mathrm{Hz}\)) to suppress high-frequency noise while preserving the dominant gait motion.

\begin{figure}[th]
  \centering
  \vspace*{3mm}
  \includegraphics[scale=0.32]{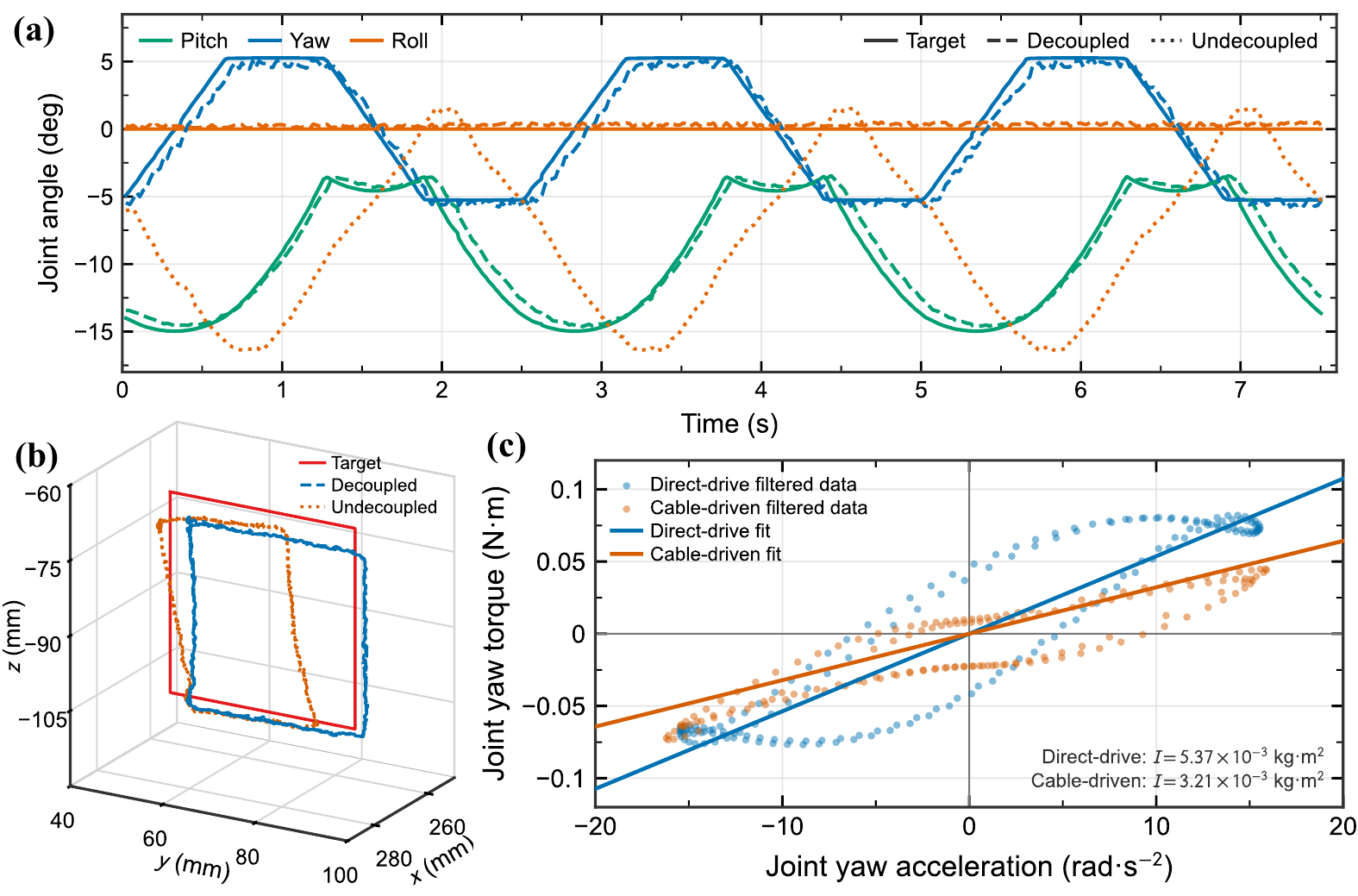}
  \vspace{0cm}
  \caption{Unloaded test of open-loop decoupling and indirect yaw-axis inertia estimation. (a) Hip angles with and without roll compensation, compared with target angles during the decoupling test. (b) Foot trajectories with and without model-based feedforward decoupling, compared with the target foot trajectory. (c) Yaw-joint torque versus angular acceleration for the cable-driven and direct-drive limbs over steady-state gait cycles, with linear-fit slopes indicating their effective rotational inertia.}
  \label{fig:decouple_inertia}
  \vspace{0cm}
\end{figure}

The effective rotational inertia was then identified as the slope of yaw-joint torque with respect to yaw-joint acceleration using linear regression, as shown in Fig. \ref{fig:decouple_inertia} (c). The identified effective yaw-axis inertia was $5.37\times10^{-3}\ \mathrm{kg\ m^2}$ for the direct-drive limb and $3.21\times10^{-3}\ \mathrm{kg\ m^2}$ for the cable-driven limb. After subtracting the estimated $6.56\times10^{-4}\ \mathrm{kg\ m^2}$ contribution of the additional force-sensor unit on the direct-drive limb, the cable-driven architecture reduced the effective yaw-axis inertia by $31.9\%$.

\subsection{Range of Motion}

To evaluate the range of motion (ROM) and transmission accuracy, we combined geometry-based workspace analysis with prototype tests over selected joint-angle ranges.

% [FK only] The reachable workspace (Fig. \ref{fig:ROM}(a)) was computed by sampling the hip pitch joint from \(-35^\circ\) to \(10^\circ\), the hip yaw joint from \(-90^\circ\) to \(35^\circ\), the hip roll joint from \(-100^\circ\) to \(100^\circ\), and the knee joint from \(-180^\circ\) to \(180^\circ\). To account for transmission coupling and mechanical interference, two constraint planes were further imposed to exclude regions inaccessible to the foot.
% The workspace volume was estimated by voxelizing the reachable foot point cloud and summing the volumes of all occupied voxels. The estimated volume was 1.28 $L^3$, where $L$ denotes the limb length.
The foot-tip workspace was estimated in Autodesk Inventor by sampling the hip yaw joint from \(-90^\circ\) to \(60^\circ\) in \(15^\circ\) increments, the hip pitch joint from \(-60^\circ\) to \(30^\circ\) in \(15^\circ\) increments, the hip roll joint from \(-90^\circ\) to \(90^\circ\) in \(15^\circ\) increments, and the knee joint from \(15^\circ\) to \(165^\circ\) in \(30^\circ\) increments. Configurations resulting in self-collision were excluded from the reachable workspace. The zero-angle configuration of all joints corresponds to the limb pose shown in Fig.~\ref{fig:overview}(b). The workspace envelope was reconstructed from the valid foot-tip positions using a regularized three-dimensional alpha shape based on Delaunay tetrahedralization \cite{edelsbrunner1994alpha}. The estimated workspace volume was approximately \(0.98~L^3\), where \(L\) denotes the limb length. Across the retained collision-free configurations, the sampled joint angles spanned \(-75^\circ\) to \(45^\circ\) for hip yaw, \(-30^\circ\) to \(0^\circ\) for hip pitch, \(-90^\circ\) to \(90^\circ\) for hip roll, and \(15^\circ\) to \(135^\circ\) for the knee. These estimates are subject to uncertainty arising from the coarse angular sampling and alpha-shape reconstruction. 
% [Manifold STL] The foot-tip workspace was estimated using collision detection of STL files  \cite{manifold} by sampling each hip joint from \(-90^\circ\) to \(90^\circ\) in \(10^\circ\) increments and the knee joint from \(-120^\circ\) to \(120^\circ\) in \(15^\circ\) increments. Configurations resulting in self-collision were excluded from the reachable workspace. The zero-angle configuration of all joints corresponds to the limb pose shown in Fig.~\ref{fig:overview}(b). The workspace envelope was reconstructed from valid foot-tip positions using a regularized three-dimensional alpha shape based on Delaunay tetrahedralization with a radius factor of 1.00 \cite{edelsbrunner1994alpha}. The smallest circumsphere-radius threshold ensuring that every sample belonged to a retained tetrahedron was selected, and the resulting boundary faces were plotted with height-based coloring and horizontal contours.

\begin{figure}[th]
  \centering
  \vspace*{3mm}
  \includegraphics[scale=0.32]{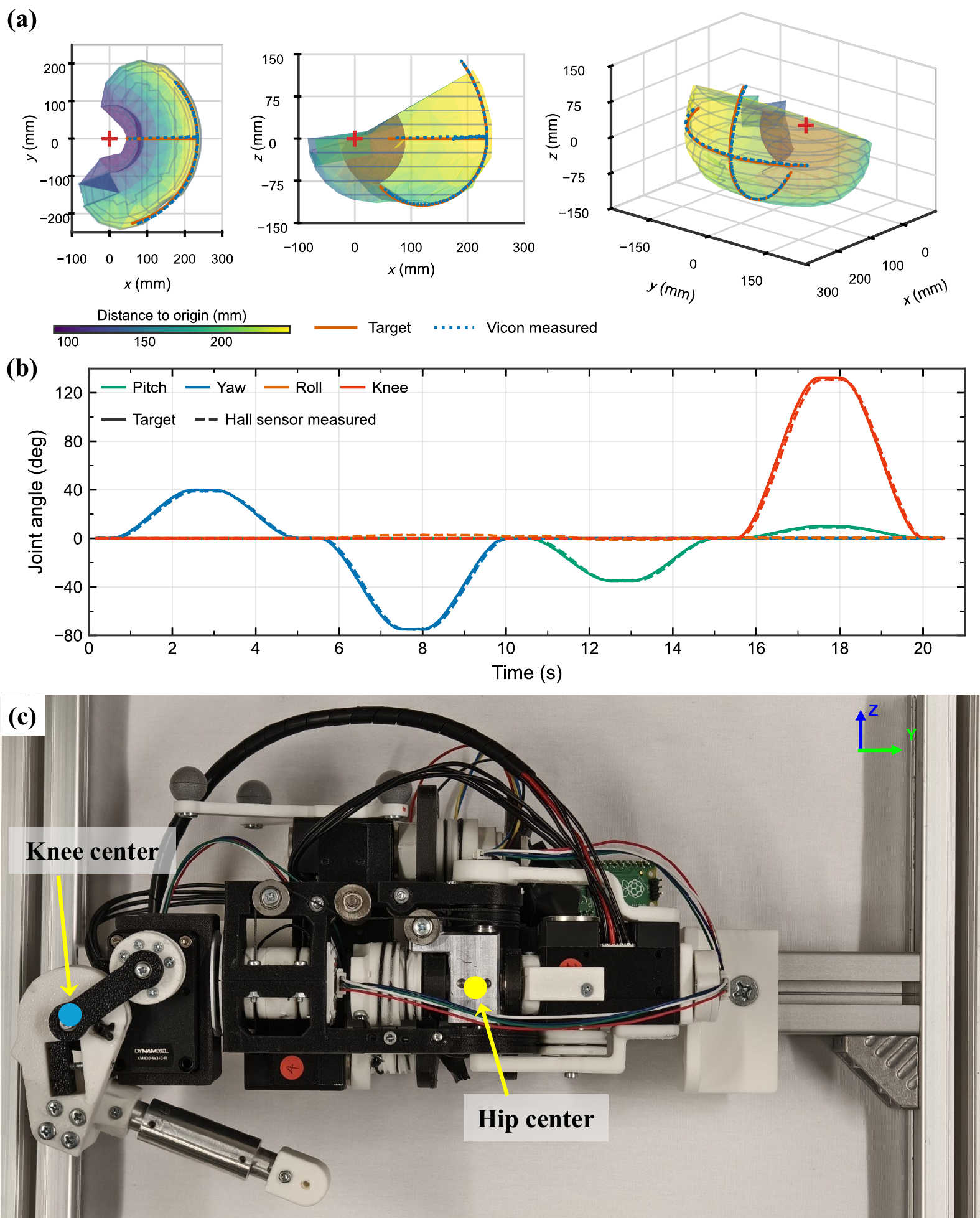}
  \vspace{0cm}
  \caption{Validation of range of motion (ROM). (a) Top, side, and oblique views of the geometry-derived foot-tip workspace estimate. Collision-free CAD samples and their reconstructed alpha-shape envelope are colored by distance from the hip center; the red cross marks the hip center. Orange solid and blue dotted curves denote the target and Vicon-measured foot trajectories, respectively. (b) Target (solid) and Hall-sensor-measured (dashed) joint angles during the joint-sweep sequence corresponding to the foot trajectories in (a). (c) Fully folded limb configuration with the yaw and knee angles manually positioned to $-90^\circ$ and $160^\circ$, respectively.}
  % \caption{Validation of unloaded motion performance: open-loop decoupling, indirect rotational-inertia comparison, and range of motion (ROM). (a) Test setup and target motions. (b) Foot trajectories with and without model-based feedforward decoupling, compared with the target foot trajectory. (c) Hip angles with and without roll compensation, compared with target angles during the decoupling test. (d) Inertia-induced yaw-motor torques of the cable-driven and direct-drive limbs in a representative gait cycle. (e-f) ROM represented by the foot trajectories (e) and joint angles (f) during the joint-sweep tests.}
  \label{fig:ROM}
  \vspace{0cm}
\end{figure}

Limitations in the current software implementation of multi-turn motor control restricted experimental testing to angular ranges of \(-35^\circ\) to \(10^\circ\), \(-78^\circ\) to \(40^\circ\), and \(0^\circ\) to \(135^\circ\) for hip pitch, hip yaw, and knee joints, respectively. 
During these tests, the hip-roll reference was set to $0^\circ$. The feedforward decoupling in Eq.~\ref{eq:ff_compensation} compensated for the roll motion induced by the other hip DoFs, maintaining the roll joint near zero (Fig.~\ref{fig:ROM}(b)). The required roll-motor compensation spanned an equivalent roll-joint motion from \(-40^\circ\) to \(78^\circ\).
% Roll motion from \(-50^\circ\) to \(48^\circ\), spanning \(98^\circ\), was additionally demonstrated during gait reproduction in the supplementary video.
Roll-angle tracking over a \(60^\circ\) range was additionally verified under load in the tests described in Section \ref{sec:load}.
The tested ranges covered the joint excursions required for the salamander walking trajectory, which is demonstrated in the supplementary video.
Output joint angles were recorded using Hall-effect sensors and compared with the angles predicted from the cable transmission ratios. These measurements were not used for feedback control in this test. Foot positions were measured with Vicon and compared with the theoretical trajectory computed from forward kinematics.

% As shown in Fig. \ref{fig:kinematics test} (e)-(f), the encoder angles follow the goal angles closely. The maximum joint-angle and foot-position deviations were $4.88^\circ$ and 8.2 mm, respectively. The resulting foot-position spans along the $x$, $y$, and $z$ axes were 181.6 mm, 377.8 mm, and 255.9 mm, and the validated yaw range of motion was $118^\circ$. 

Both the measured foot trajectories and joint angles closely match the target ones.
The foot-position spans attained during the tests along the $x$, $y$, and $z$ axes were 78.9\%, 164.3\%, and 111.3\% of the limb length, respectively (Fig. \ref{fig:ROM} (a)). The mean three-dimensional foot-position error was 4.78 mm, corresponding to 2.08\% of the 230-mm limb length.
The encoder-recorded output-side joint angles follow the goal angles closely (Fig. \ref{fig:ROM} (b)). The roll joint, which has the longest transmission path, exhibits a mean absolute tracking error (MAE) of \(0.91^\circ\).
Manual positioning additionally confirmed that the prototype could adopt a fully folded posture along the trunk for swimming (Fig. \ref{fig:ROM}(d)). Hip yaw and knee angles of \(-90^\circ\) and \(160^\circ\), respectively, were reached manually. This demonstrates the physical attainability of the folded posture, but motor-driven operation at these angles remains unverified. Future work will address the multi-turn control limitation and extend experimental validation toward the full predicted geometric ROM.

\subsection{Load Test} \label{sec:load}

To validate whether the new design could generate sufficient joint torques for locomotion, we subjected the limb to loads in multiple directions using the two platforms shown in Fig. \ref{fig:experiment_platforms} (b)-(c) and compared the results with those from the direct-drive limb subject to the same load (Fig. \ref{fig:experiment_platforms} (d)-(e)). We used weights of 1.25 and 2.5~kg to apply loads. During the load tests, the cable-driven limb was controlled using the model-based feedforward decoupling described in Section \ref{sec:decoupling}. Output-side joint angle feedback control was only used in the yaw and roll joint load tests. Motor torques were estimated from motor currents as in Section \ref{sec:exp_inertia}, but without scaling by the transmission reduction ratios so that the values directly reflect motor loading. The torque signals were sampled at 50 Hz without any filtering.

For the pitch and knee joints, weights were mounted on a four-bar linkage \cite{Birdbot} that maintained the trunk platform orientation while allowing vertical translation. The ball foot was replaced with rollers to avoid imposing unintended horizontal constraints. The moving platform components imposed an additional load exceeding 1.8~kg, depending on the angles between the links of the four-bar linkage.
The pitch and knee trajectories were generated through inverse kinematics by prescribing a 5.0-s periodic circular-arc motion of the hip center while keeping the foot stationary. The angular ranges of the pitch and knee joints were \(10.39^\circ\) and \(37.71^\circ\), respectively, while their peak angular velocities were \(10.44^\circ\,\mathrm{s}^{-1}\) and \(28.00^\circ\,\mathrm{s}^{-1}\).

For the yaw and roll joints, the weight was connected to the foot through a cable, and a redirecting pulley converted the vertical gravitational load into a horizontal cable force acting on the tested joint. 
The yaw and roll joints were controlled to follow sinusoidal trajectories over a \(60^\circ\) range with a period of 7.5 s and a peak angular velocity of \(25.13^\circ\,\mathrm{s}^{-1}\).

% \begin{figure}[th]
%   \centering
%   \vspace*{2mm}
%   \includegraphics[scale=0.95]{Fig/Fig-load platform.png}
%   \vspace{0cm}
%   \caption{Experimental setups for joint load tests. }
%   \label{fig:load test platform}
%   \vspace{0cm}
% \end{figure}

\begin{figure}[th]
  \centering
  \vspace*{2mm}
  \includegraphics[scale=0.32]{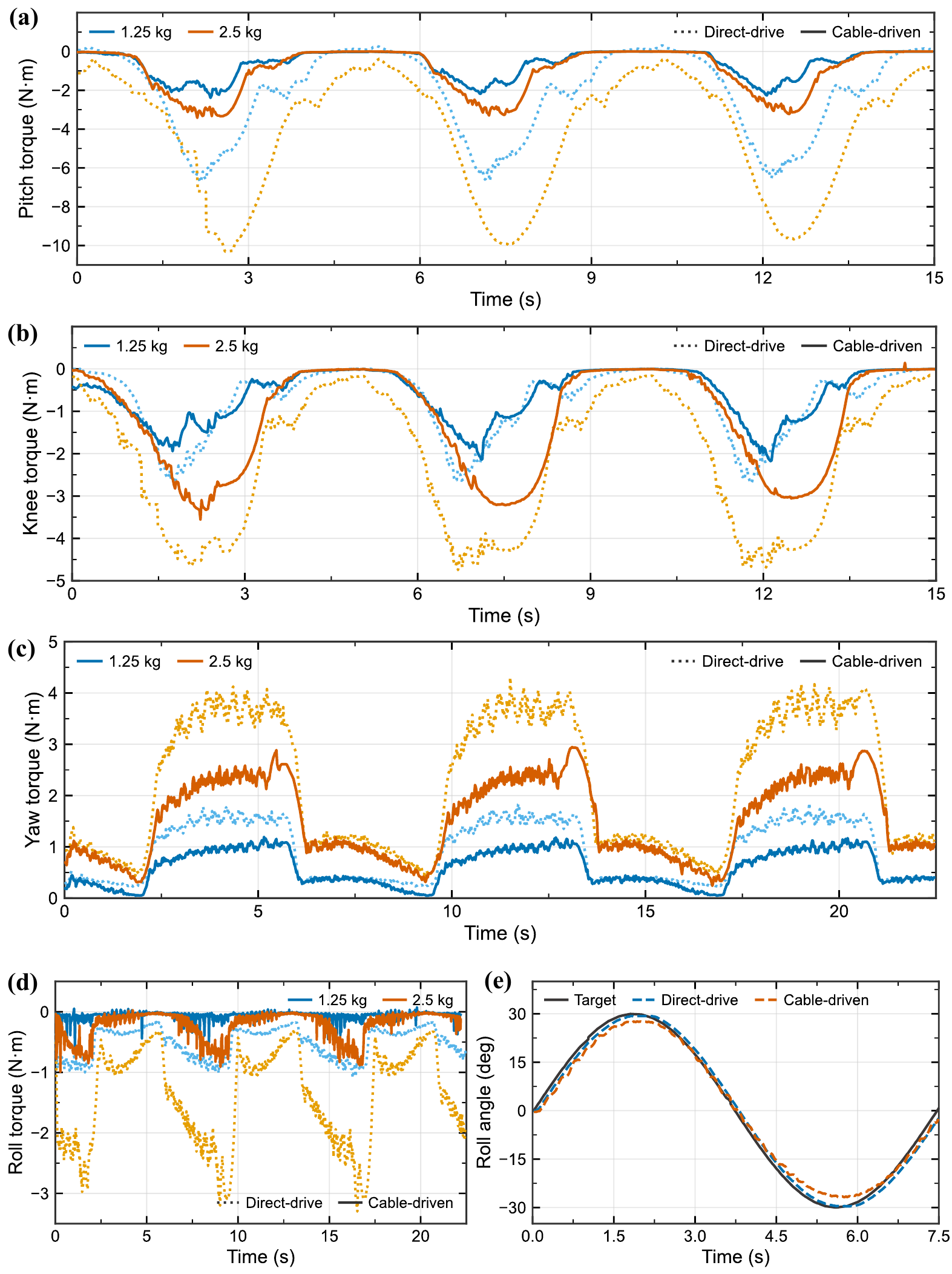}
  \vspace{0cm}
  \caption{Load-test comparison of the cable-driven (solid) and direct-drive (dash-dotted) limbs. Motor-torque histories for (a) hip pitch, (b) knee, (c) hip yaw, and (d) hip roll with added masses of 1.25 and 2.5 kg. (e) Roll-angle reference and measured responses of the two limbs in one representative cycle.}
  \label{fig:load test data}
  \vspace{0cm}
\end{figure}

% Fig. \ref{fig:load test data} (a) and (b) show that, under matched loading conditions, the cable-driven limb required lower pitch and knee motor torques than the direct-drive limb. %The cable-driven limb raised a maximum load of 2.5 kg. At this load, the pitch joint delivered approximately 10 N·m, the highest measured joint torque, while the knee joint delivered approximately 4.2 N·m. The yaw test further demonstrated a maximum external load of 2.5 kg and a joint torque of approximately 3.75 N·m.
The results demonstrated that our design reduces motor torque demand through its reduction ratio while still meeting the speed requirements of locomotion \cite{Pleurobot}. Supporting an equivalent vertical load corresponding to a mass of at least 4.3 kg, including the weight and the moving platform, the pitch and knee motors of the direct-drive limb required peak torque magnitudes of 10.3 and 4.7~N·m, respectively. Under the same motion and external load, the cable-driven limb required only 3.4~N·m at the pitch joint and 3.6~N·m at the knee joint, corresponding to reductions of 67.0\% and 23.4\%, respectively. Similar reductions were observed at the hip yaw and roll joints (Fig.~\ref{fig:load test data}).

For the roll joint, with an additional 2.5-kg mass generating a horizontal force through the redirecting pulley, the peak motor torque magnitude in the cable-driven leg was 0.99~N·m, 30.3\% of that required by the direct-drive leg (Fig. \ref{fig:load test data} (d)). With combined feedforward decoupling and feedback control, the maximum roll-angle tracking error was $5.50^\circ$ for the cable-driven limb, larger than the $2.02^\circ$ for the direct-drive limb (Fig. \ref{fig:load test data} (e)). The larger error is consistent with the lower stiffness associated with the longer cable path. 

%a maximum measured joint torque of approximately 10 N·m and enabling the limb to raise a 5.8-kg external load.

%% file: Section/05-Conclusion.tex
\section{Conclusion}
The results demonstrate that the proposed cable-driven limb reduces mass, size, and estimated effective yaw-axis inertia relative to the direct-drive limb while retaining large-amplitude, 4-DoF salamander-like limb motions and high payload capacity. The mass reduction is achieved primarily by optimizing the transmission reduction ratios, enabling the use of smaller motors, while the reduced estimated effective yaw-axis inertia and size result from relocating heavier motors proximally. These benefits are enabled by mechanically compensating multi-segment cable paths and model-based feedforward decoupling, which address transmission-induced coupling while preserving the required ROM and load capacity. Output-side joint-angle feedback further improves tracking accuracy under load. 
% The results demonstrate that the proposed cable-driven limb reduces mass, lateral cross-sectional area, and effective yaw-axis rotational inertia relative to the direct-drive limb while retaining 4-DoF salamander-like motion and the required load capacity. Spatial cable routing accommodates the transmissions within a compact hip while preserving three mutually orthogonal axes intersecting at a common center, and the integrated pretensioning mechanisms avoid additional standalone assemblies. Joint-specific reduction ratios enable smaller motors to satisfy the joint torque and ROM requirements. The routing mechanically eliminates total cable-length changes, while model-based feedforward compensation addresses the differential cable-length changes induced by pitch and yaw rotations. Output-side joint-angle feedback further improves tracking under load.

However, the cable-driven roll joint exhibited larger tracking errors than its direct-drive counterpart, presumably because of elastic deformation along the longer cable path. In addition, the complex cable routing and pretensioning mechanisms increase maintenance difficulty. These findings highlight the trade-off between these reductions enabled by cable transmission and the resulting decrease in transmission stiffness and maintainability. Future work will focus on improving distal transmission stiffness and control performance, modularizing the transmission to facilitate maintenance, and integrating the limb into a complete robot for locomotion experiments.

%% file: ref.bib
@article{understanding_salamander_locomotion,
  title={Where are we in understanding salamander locomotion: biological and robotic perspectives on kinematics},
  author={Karakasiliotis, Konstantinos and Schilling, Nadja and Cabelguen, Jean-Marie and Ijspeert, Auke Jan},
  journal={Biological cybernetics},
  volume={107},
  number={5},
  pages={529--544},
  year={2013},
  publisher={Springer}
}

@inproceedings{D3-arm,
  title={{D3-Arm}: High-Dynamic, Dexterous and Fully Decoupled Cable-driven Robotic Arm},
  author={Luo, Hong and Xu, Jianle and Li, Shoujie and Liang, Huayue and Chen, Yanbo and Xia, Chongkun and Wang, Xueqian},
  booktitle={2025 IEEE International Conference on Robotics and Automation (ICRA)},
  pages={6252--6258},
  year={2025},
  organization={IEEE}
}

@article{haoqileg,
  title={High-Degree-of-Freedom Lightweight Bioinspired Leg for Enhanced Mobility in Small Robots},
  author={Han, Haoqi and Yu, Yifei and Zhang, Jiaming and Cui, Xinru and Feng, Linxi and Wang, Hesheng},
  journal={arXiv preprint arXiv:2606.18680},
  year={2026}
}

@article{liuCPG,
  title={A novel design methodology of {CPG} model for a salamander-like robot},
  author={Liu, Yang and Liu, Zhiang and Fang, Yongchun and Liu, Huawang and Guo, Xian},
  journal={IEEE Robotics and Automation Letters},
  volume={9},
  number={7},
  pages={6115--6122},
  year={2024},
  publisher={IEEE}
}

@inproceedings{TwistSnake,
  title={Twist Snake: Plastic table-top cable-driven robotic arm with all motors located at the base link},
  author={Tanaka, Kazutoshi and Hamaya, Masashi},
  booktitle={2023 IEEE International Conference on Robotics and Automation (ICRA)},
  pages={7345--7351},
  year={2023},
  organization={IEEE}
}

@article{Pleurobot,
    author = {Karakasiliotis, K. and Thandiackal, R. and Melo, K. and Horvat, T. and Mahabadi, N. K. and Tsitkov, S. and Cabelguen, J. M. and Ijspeert, A. J.},
    title = {From cineradiography to biorobots: an approach for designing robots to emulate and study animal locomotion},
    journal = {Journal of The Royal Society Interface},
    volume = {13},
    number = {119},
    pages = {20151089},
    year = {2016},
    month = {06},
    issn = {1742-5689},
    doi = {10.1098/rsif.2015.1089},
    url = {https://doi.org/10.1098/rsif.2015.1089},
}

@ARTICLE{Auke13TRO,
  author={Crespi, Alessandro and Karakasiliotis, Konstantinos and Guignard, André and Ijspeert, Auke Jan},
  journal={IEEE Transactions on Robotics}, 
  title={Salamandra Robotica {II}: An Amphibious Robot to Study Salamander-Like Swimming and Walking Gaits}, 
  year={2013},
  volume={29},
  number={2},
  pages={308-320},
  doi={10.1109/TRO.2012.2234311}}

@article{
leg_review,
author = {Junhui Zhang  and Jinyuan Liu  and Huaizhi Zong  and Pengyuan Ji  and Lizhou Fang  and Yong Li  and Huayong Yang  and Bing Xu },
title = {Bridging the Gap to Bionic Motion: Challenges in Legged Robot Limb Unit Design, Modeling, and Control},
journal = {Cyborg and Bionic Systems},
volume = {6},
number = {},
pages = {0365},
year = {2025},
doi = {10.34133/cbsystems.0365},
URL = {https://spj.science.org/doi/abs/10.34133/cbsystems.0365},
eprint = {https://spj.science.org/doi/pdf/10.34133/cbsystems.0365}}

@ARTICLE{Cable_Arm_Kim,
  author={Kim, Yong-Jae},
  journal={IEEE Transactions on Robotics}, 
  title={Anthropomorphic Low-Inertia High-Stiffness Manipulator for High-Speed Safe Interaction}, 
  year={2017},
  volume={33},
  number={6},
  pages={1358-1374},
  doi={10.1109/TRO.2017.2732354}}

@misc{xm430w350emanual,
  author = {{ROBOTIS}},
  title = {{XM430-W350}},
  url = {https://docs.robotis.com/docs/dxl/model_reference/x_series/xm_series/xm430-w350/},
  note = {Accessed: 2026-09-14}
}

@misc{Hand_haoqi,
      title={Development of a 15-Degree-of-Freedom Bionic Hand with Cable-Driven Transmission and Distributed Actuation}, 
      author={Haoqi Han and Yi Yang and Yifei Yu and Yixuan Zhou and Xiaohan Zhu and Hesheng Wang},
      year={2025},
      eprint={2512.04399},
      archivePrefix={arXiv},
      primaryClass={cs.RO},
      url={https://arxiv.org/abs/2512.04399}, 
}

@article{hand_spring,
  title={An 18-{DOF} hand integrating force--position multimodal perception using a monocular camera},
  author={Chen, Shiwei and Li, Jiapeng and Deng, Zhiming and Wang, Peiji and Wei, Cheng and Cao, Xibin},
  journal={Nature Communications},
  volume={16},
  number={1},
  pages={6801},
  year={2025},
  publisher={Nature Publishing Group UK London}
}

@INPROCEEDINGS{cable_driven_dog_leg,
  author={Hwangbo, Jemin and Tsounis, Vassilios and Kolvenbach, Hendrik and Hutter, Marco},
  booktitle={2018 IEEE/RSJ International Conference on Intelligent Robots and Systems (IROS)}, 
  title={Cable-Driven Actuation for Highly Dynamic Robotic Systems}, 
  year={2018},
  volume={},
  number={},
  pages={8543-8550},
  doi={10.1109/IROS.2018.8593569}}

@article{Birdbot,
author = {Alexander Badri-Spröwitz  and Alborz Aghamaleki Sarvestani  and Metin Sitti  and Monica A. Daley },
title = {{BirdBot} achieves energy-efficient gait with minimal control using avian-inspired leg clutching},
journal = {Science Robotics},
volume = {7},
number = {64},
pages = {eabg4055},
year = {2022},
doi = {10.1126/scirobotics.abg4055},
URL = {https://www.science.org/doi/abs/10.1126/scirobotics.abg4055},
eprint = {https://www.science.org/doi/pdf/10.1126/scirobotics.abg4055}
}

@article{Low_Cost_Cable-Driven_Robot_Arm,
AUTHOR = {Nguyen, Van Pho and Chow, Wai Tuck and Dhyan, Sunil Bohra and Zhang, Bohan and Han, Boon Siew and Wong, Hong Yee Alvin},
TITLE = {Low-Cost Cable-Driven Robot Arm with Low-Inertia Movement and Long-Term Cable Durability},
JOURNAL = {Robotics},
VOLUME = {13},
YEAR = {2024},
NUMBER = {9},
pages = {128},
URL = {https://www.mdpi.com/2218-6581/13/9/128},
ISSN = {2218-6581},
DOI = {10.3390/robotics13090128}
}

@article{TITAN-XIII,
  title={{TITAN-XIII}: sprawling-type quadruped robot with ability of fast and energy-efficient walking},
  author={Kitano, Satoshi and Hirose, Shigeo and Horigome, Atsushi and Endo, Gen},
  journal={Robomech Journal},
  volume={3},
  number={1},
  pages={8},
  year={2016},
  publisher={Springer}
}

@INPROCEEDINGS{Constant_Length_Tendon,
  author={Shah, Divya and Parmiggiani, Alberto and Kim, Yong-Jae},
  booktitle={2020 IEEE/ASME International Conference on Advanced Intelligent Mechatronics (AIM)}, 
  title={Constant Length Tendon Routing Mechanism through Axial Joint}, 
  year={2020},
  volume={},
  number={},
  pages={753-758},
  doi={10.1109/AIM43001.2020.9158889}}

@INPROCEEDINGS{decoupling_four_link,
  author={Lee, Jong Kwang and Choi, Chang Hwan and Yoon, Kwang Ho and Lee, Hyo Jik and Park, Byung Suk and Yoon, Ji Sup},
  booktitle={2008 International Conference on Smart Manufacturing Application}, 
  title={Design and Evaluation of Cable-driven Manipulator with Motion-decoupled Joints}, 
  year={2008},
  volume={},
  number={},
  pages={575-580},
  doi={10.1109/ICSMA.2008.4505591}}

@article{decoupling_rotary_pulley,
  title={Design and modeling of motion-decoupling mechanism for cable-driven joints},
  author={Jiang, Surong and Hua, Daren and Wang, Yaoyao and Ju, Feng and Yin, Liang and Chen, Bai},
  journal={Advances in Mechanical Engineering},
  volume={10},
  number={5},
  pages={1687814018777428},
  year={2018},
  publisher={SAGE Publications Sage UK: London, England}
}

@article{decouple_movable_pulley,
title = {Design and implementation of a 7-{DOF} cable-driven serial spray-painting robot with motion-decoupling mechanisms},
journal = {Mechanism and Machine Theory},
volume = {192},
pages = {105549},
year = {2024},
issn = {0094-114X},
doi = {10.1016/j.mechmachtheory.2023.105549},
url = {https://www.sciencedirect.com/science/article/pii/S0094114X23003208},
author = {Feng Xu and Bin Zi and Zhaoyi Yu and Jiahao Zhao and Huafeng Ding}
}

@ARTICLE{decouple_Geared_Rolling_Joints,
  author={Choi, Wontae and Suh, Jungwook},
  journal={IEEE/ASME Transactions on Mechatronics}, 
  title={Design and Evaluation of a Cable-Actuated Palletizing Robot With Geared Rolling Joints}, 
  year={2024},
  volume={29},
  number={4},
  pages={3137-3145},
  doi={10.1109/TMECH.2024.3403131}}

@INPROCEEDINGS{ORCA_Hand,
  author={Christoph, Clemens C. and Eberlein, Maximilian and Katsimalis, Filippos and Roberti, Arturo and Sympetheros, Aristotelis and Vogt, Michel R. and Liconti, Davide and Yang, Chenyu and Cangan, Barnabas Gavin and Hinchet, Ronan J. and Katzschmann, Robert K.},
  booktitle={2025 IEEE/RSJ International Conference on Intelligent Robots and Systems (IROS)}, 
  title={{ORCA}: An Open-Source, Reliable, Cost-Effective, Anthropomorphic Robotic Hand for Uninterrupted Dexterous Task Learning}, 
  year={2025},
  volume={},
  number={},
  pages={8503-8510},
  doi={10.1109/IROS60139.2025.11246681}
}

@article{differential_dog_leg,
  title={A Differential-Mechanism-Based Leg Configuration Balances the Load and Dynamic Contribution for All Actuators of the Quadruped Robot},
  author={Wang, Zeyu and Jia, Wenchuan and Sun, Yi and Bao, Tianxu and Ding, Zihan and Chen, Qi},
  journal={IEEE Transactions on Robotics},
  volume={41},
  pages={2014--2030},
  year={2025},
  publisher={IEEE}
}

@inproceedings{Stanford_doggo,
  title={Stanford doggo: An open-source, quasi-direct-drive quadruped},
  author={Kau, Nathan and Schultz, Aaron and Ferrante, Natalie and Slade, Patrick},
  booktitle={2019 International conference on robotics and automation (ICRA)},
  pages={6309--6315},
  year={2019},
  organization={IEEE}
}

@article{auke_amphibious,
  title={Amphibious and sprawling locomotion: from biology to robotics and back},
  author={Ijspeert, Auke J},
  journal={Annual Review of Control, Robotics, and Autonomous Systems},
  volume={3},
  number={1},
  pages={173--193},
  year={2020},
  publisher={Annual Reviews}
}

@phdthesis{pleurobot_phd_thesis,
  title={Legged locomotion with spinal undulations},
  author={Karakasiliotis, Konstantinos},
  year={2013},
  school={EPFL}
}

@inproceedings{rolling_hand,
  title={Getting the ball rolling: Learning a dexterous policy for a biomimetic tendon-driven hand with rolling contact joints},
  author={Toshimitsu, Yasunori and Forrai, Benedek and Cangan, Barnabas Gavin and Steger, Ulrich and Knecht, Manuel and Weirich, Stefan and Katzschmann, Robert K},
  booktitle={2023 IEEE-RAS 22nd International Conference on Humanoid Robots (Humanoids)},
  pages={1--7},
  year={2023},
  organization={IEEE}
}

@article{kawano_blob_2022,
    author = {Kawano, Sandy M. and Blob, Richard W.},
    title = {Terrestrial force production by the limbs of a semi-aquatic salamander provides insight into the evolution of terrestrial locomotor mechanics},
    journal = {Journal of Experimental Biology},
    volume = {225},
    number = {7},
    pages = {jeb242795},
    year = {2022},
    month = {04},
    issn = {0022-0949},
    doi = {10.1242/jeb.242795},
    url = {https://doi.org/10.1242/jeb.242795},
    eprint = {https://journals.biologists.com/jeb/article-pdf/225/7/jeb242795/2139247/jeb242795.pdf},
}

@article{edelsbrunner1994alpha,
  author  = {Edelsbrunner, Herbert and M{\"u}cke, Ernst P.},
  title   = {Three-Dimensional Alpha Shapes},
  journal = {ACM Transactions on Graphics},
  year    = {1994},
  volume  = {13},
  number  = {1},
  pages   = {43--72},
  doi     = {10.1145/174462.156635}
}
